\documentclass[runningheads]{llncs}

\usepackage{eccv}

\usepackage{eccvabbrv}

\usepackage{graphicx}
\usepackage{booktabs}
\usepackage{array, makecell}

\usepackage[accsupp]{axessibility}  

\usepackage{hyperref}

\usepackage{orcidlink}

\begin{document}

\title{SceneTeller: Language-to-3D Scene Generation} 

\titlerunning{SceneTeller: Language-to-3D Scene Generation}

\author{Başak Melis Öcal\inst{1}\thanks{Correspondence to <\email{b.m.ocal@uva.nl}>} \and
Maxim Tatarchenko\inst{2} \and
Sezer Karaoğlu\inst{1} \and 
Theo Gevers\inst{1}}

\authorrunning{B.M. Öcal et al.}

\institute{UvA-Bosch Delta Lab, University of Amsterdam \and
Bosch Center for AI, Robert Bosch GmbH
}

\maketitle

\begin{figure}[h]
  \centering
   \includegraphics[height=10.9cm]{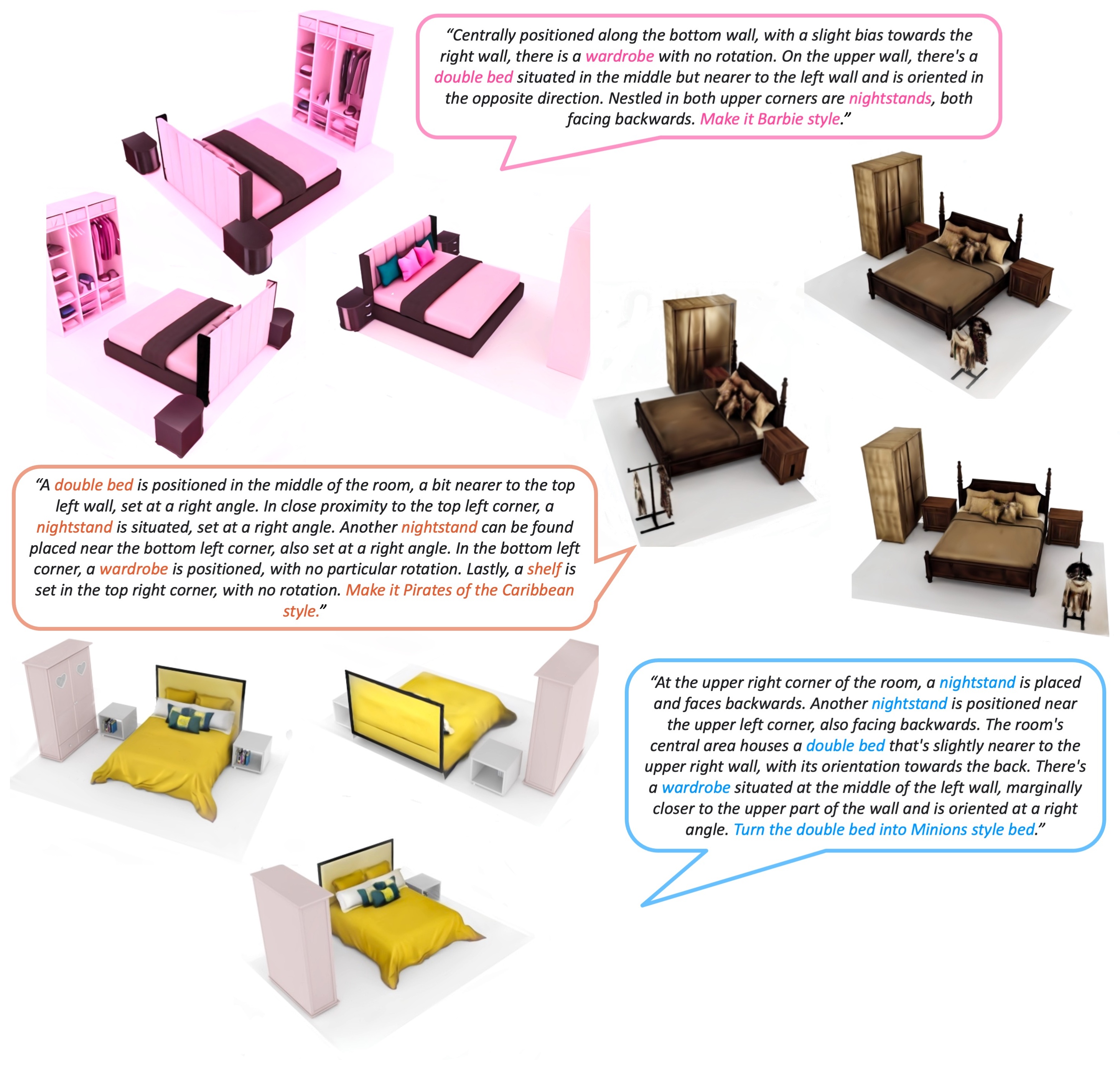}
   \caption{With solely a textual prompt in natural language describing the desired spatial positions and orientations of objects within the scene, \textit{SceneTeller} is able to generate realistic and high-quality 3D spaces. Additionally, by facilitating modifications to the style of the entire scene or individual objects within the scene through edit instructions, \textit{SceneTeller} offers a practical and flexible framework for designing personalized rooms.}
   \label{fig:introqual}
\end{figure}

\begin{abstract}

Designing high-quality indoor 3D scenes is important in many practical applications, such as room planning or game development. Conventionally, this has been a time-consuming process which requires both artistic skill and familiarity with professional software, making it hardly accessible for layman users. However, recent advances in generative AI have established solid foundation for democratizing 3D design. In this paper, we propose a pioneering approach for text-based 3D room design. Given a prompt in natural language describing the object placement in the room, our method produces a high-quality 3D scene corresponding to it. With an additional text prompt the users can change the appearance of the entire scene or of individual objects in it. Built using in-context learning, CAD model retrieval and 3D-Gaussian-Splatting-based stylization, our turnkey pipeline produces state-of-the-art 3D scenes, while being easy to use even for novices. Our project page is available at \href{https://sceneteller.github.io/}{https://sceneteller.github.io/}.
  
\end{abstract}

\section{Introduction}
\label{sec:intro}

3D scene design has numerous applications in domains like architecture or game development. Designing high-quality 3D scenes, however, is a laborious and time-consuming process, challenging even for highly-skilled domain experts utilizing professional tools. This complexity often leaves novice users unable to customize their own spaces. Recent advancements in automatic text-to-image synthesis \cite{ddpm, stablediffusion} have led to massive progress in text-to-3D creation \cite{dreamfusion, dreambooth3d, magic123}, enabling the generation of 3D content only from text input. Such works create foundation for democratizing 3D design, supporting its accessibility for non-experts.

A pioneering work by Poole et al.~\cite{dreamfusion} introduced object-centric text-to-3D generation based on 2D diffusion models coupled with NeRFs \cite{mildenhall2020nerf}.
Several subsequent works proposed ways to improve the geometric accuracy, multi-view consistency, and appearance quality \cite{prolificdreamer, magic3d, fantasia3d, latentnerf}.
Despite their success in object-level generation, hallucinating complex large-scale scenes with multiple objects is still remaining a challenging problem. Some recent 3D scene generation methods \cite{setthescene,comp3d,discoscene} require a 3D layout as input, limiting their usability for novices. Others \cite{scenescape, text2room} produce results that violate global object arrangements and are not multi-view consistent.

Aiming to address both these shortcomings - insufficient user friendliness and unsatisfactory global consistency - we introduce \textit{SceneTeller}, a framework for text-based 3D scene generation. 1) We start with a premise that a 3D scene generation framework should offer control over the placement and style of individual objects, allowing novice users to design their customized spaces through simple interactions with the system. Our approach achieves this by relying on text-based user interaction for scene and appearance specification. 2) In order to generate compositionally plausible 3D scenes, it is important to incorporate some degree of global information guidance. We implement this by conditioning the method on global layout descriptions.

To this end, we split the task of generating high-quality 3D scenes into three steps: 3D layout generation, object population and style editing. 1) When presented with a prompt describing the desired spatial positions and orientations of objects, \textit{SceneTeller} generates an initial 3D layout of the scene by exploiting the Large Language Models (LLMs) through in-context learning, to ensure global consistency. 2) The generated layout is then used to assemble a 3D scene and populate it with CAD models from a database. 3) To facilitate scene editing, the assembled scene is fitted with a 3D Gaussian Splatting \cite{gaussiansplatting} representation. This representation is then used to stylize the scene according to the user-provided edit prompt targeting the entire scene or individual objects within it.

Extensive quantitative and qualitative evaluations show that \textit{SceneTeller} performs significantly better than state-of-the-art methods, offering superior geometric fidelity and compositional plausibility. In brief, our contributions to the existing body of work can be summarized as follows:

\begin{itemize}

\item We introduce a turnkey pipeline for high-quality 3D scene generation yielding complex, realistic and globally consistent 3D scenes, given only scene descriptions in natural language. Our pipeline supports object- and scene-level appearance editing.

\item We propose a novel LLM-based module for 3D layout generation trained through in-context learning, which provides control over individual object placement.



\item We experimentally demonstrate that \textit{SceneTeller} generates 3D scenes of higher quality and better consistency than state-of-the-art methods, while boasting unprecedented user friendliness and flexibility.

\end{itemize}

\section{Related Work}
\label{sec:relatedwork}

\subsection{3D Scene Representations}
\label{subsec:related3dscenerep}

Rendering 3D scenes using implicit neural networks has seen remarkable success, notably with NeRF's \cite{mildenhall2020nerf} photo-realistic image synthesis. Extensions of NeRF improve reconstruction quality \cite{uhdnerf, mipnerf}, training and rendering speed \cite{plenoxels, plenoctrees, merf, directvoxelgrid,tensorf,instantneuralgraphics}, and handling large-scale scenes \cite{hallucinatednerf, blocknerf}. Despite advancements, NeRF-based methods still struggle with slow rendering speeds. Recently, 3D Gaussian splatting (3DGS) was proposed for real-time, high-quality rendering using anisotropic 3D Gaussians and fast tile-based rasterization. Our framework employs 3DGS for scene representation and allows edits on these Gaussians, generating high-quality, 3D consistent scenes.


\subsection{Text-to-3D Generation}
\label{subsec:relatedtexto3dgen}

\subsubsection{Object-centric Generation.} 

3D generation from textual prompts has recently grown in popularity. DreamFusion \cite{dreamfusion} introduces Score Distillation Sampling (SDS) to optimize NeRF representations by leveraging an off-the-shelf 2D diffusion model, paving the way towards text-driven 3D content creation \cite{dreambooth3d, att3d, hifa, magic123}. Following \cite{dreamfusion}, Magic3D \cite{magic3d} employs a coarse-to-fine optimization scheme, exploiting low-resolution diffusion priors, and then enhancing the texture. To achieve high-fidelity mesh generation, Fantasia3D \cite{fantasia3d} disentangles the appearance from geometry. Latent-NeRF \cite{latentnerf} optimizes a NeRF in the latent space of StableDiffusion \cite{stablediffusion}, and additionally proposes incorporating 3D priors to generate a target geometry. To address the over-smoothing, over-saturation and lack of diversity problems within the SDS, ProfilicDreamer \cite{prolificdreamer} proposes a principled particle-based variational framework, namely Variational Score Distillation (VSD). Following the same motivation, \cite{sjc} presents Perturb-and-Average Scoring to estimate the score for non-noisy images. 

With the recent introduction of 3D Gaussian Splatting \cite{gaussiansplatting}, several works attempt to combine 3DGS and diffusion models. GaussianDreamer \cite{gaussiandreamer} employs a 3D diffusion model to generate the initialized point clouds for 3D Gaussians, and then further optimize the Gaussians using SDS with a 2D diffusion model. \cite{gsgen,dreamgaussian} follows a two-stage pipeline for geometry optimization and appearance refinement. GaussianDiffusion \cite{gaussiandiffusion} focuses on the challenge of achieving multi-view consistency in 3D generation and proposes combining structured noise with a variational 3DGS technique. Despite their success in generating object-level scenes, both NeRF-based and 3DGS-based methods fall short in hallucinating complex scenes with multiple objects. 

\subsubsection{Scene Generation.}

To represent multiple objects within the scenes, existing works \cite{setthescene, componerf, comp3d, discoscene} rely on compositional NeRFs that are initialized based on user-defined object proxies. However, they are ususally restricted to produce only narrow inward-facing views of the rooms. Also, the requirement of having 3D layouts as input is restrictive for final users. Orthogonal to those, SceneScape \cite{scenescape} and Text2Room \cite{text2room} generate scenes following the \textit{render-refine-repeat} paradigm by iteratively warping and inpainting the previously generated image according to the specified camera motion. This strategy does not allow object-level control during scene generation, and the results tend to exhibit multi-view inconsistencies. LucidDreamer \cite{luciddreamer} employs the same paradigm for an initial point cloud reconstruction, and then further optimizes the scene using 3DGS. Recently, several works introduced techniques for 3DGS scene editing, by manipulating the training images using 2D diffusion models \cite{igs2gs, gaussianeditor1, gaussianeditor2} and incorporating the edited images into 3DGS scenes for further optimization. Concurrent to our work, \cite{gala3d} initializes a 3DGS scene from the layouts created using LLMs and optimizes it using a diffusion model. However, \cite{gala3d} lacks control over the generated layout in terms of positions and orientations of objects.

\subsection{LLMs for Vision}
\label{subsec:texto3dgen}

With the inclusion of grounding information in LLMs, they demonstrate enhanced reasoning capabilities for visual concepts. Recently, the visual reasoning abilities of LLMs have been utilized for 3D tasks such as avatar generation \cite{avatarsimulation}, instruction-driven 3D modeling \cite{3dgpt}, 3D editing \cite{gaussianeditor}. Exploiting the spatial reasoning capabilities of LLMs, LayoutGPT \cite{layoutgpt} extends beyond its 2D counterparts \cite{groundedtextoimage} to generate random 3D layouts of specific room types through in-context learning. However, LayoutGPT lacks support for controlling the generated layout. The 3D layout generation module of \textit{SceneTeller} is the first of its kind allowing control over individual object placement in accordance with the provided natural language prompts describing the desired positions of objects.

\section{Method}
\label{sec:method}

\begin{figure}[h]
  \centering
   \includegraphics[height=5.7cm]{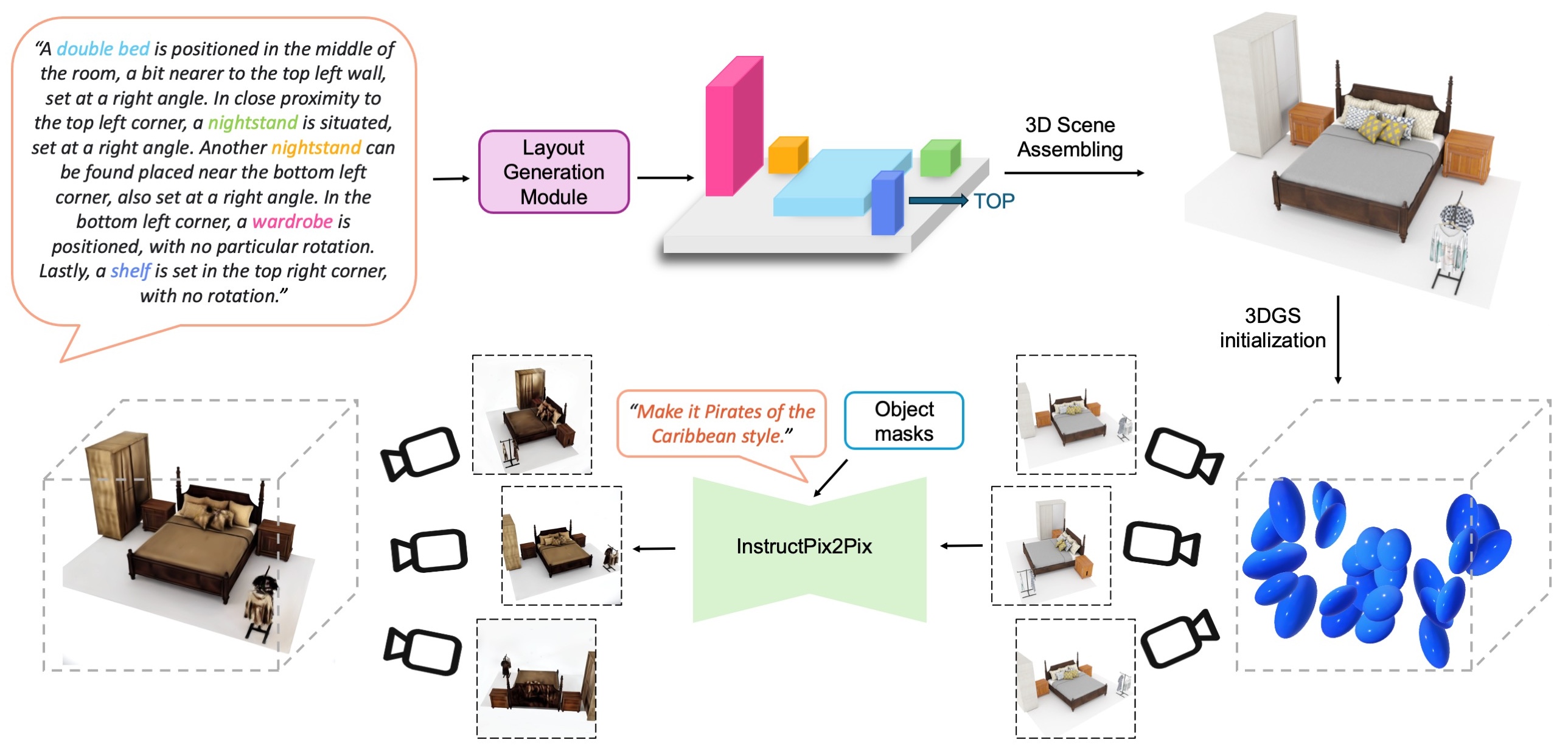}
   \caption{\textbf{Overview of our method.} Given a textual prompt in natural language delineating the desired spatial positions and orientations of objects within the scene, a 3D scene layout is generated using in-context learning. An initial 3D scene is assembled for the predicted layout, which is then fitted with a 3D Gaussian Splatting representation. This representation is then used to stylize the scene according to the user-provided text prompt, and subsequently render the final images of the scene.}
   \label{fig:arch}
\end{figure}

The pipeline of \textit{SceneTeller} is divided into three main stages: language-driven 3D layout generation, 3D scene assembling from the layout and 3D scene stylization. In the first stage, a 3D scene layout is generated based on the given text prompt (\cref{subsec:languagedriven3dlayoutgeneration}). Then, an initial 3D scene is assembled for the predicted layout (\cref{subsec:3dsceneassemblingfromlayouts}). After that, a 3D Gaussian Splatting representation is fitted to the assembled scene. This representation is then used to stylize the scene according to the user-provided text prompt, and subsequently render the final images of the scene (\cref{subsec:3dscenestylization}. The scheme of the framework is illustrated in \cref{fig:arch}

\subsection{Language-driven 3D Layout Generation}
\label{subsec:languagedriven3dlayoutgeneration}

The task of this module is to produce a collection of object bounding boxes conditioned on a brief textural description of their spatial positions within the scene in plain language.
More formally, the input is a condition $\mathcal{C}$, consisting of a textual description $\mathcal{Y}$, along with the scene type and scene dimensions information. The output is a 3D layout represented as a set of 3D bounding boxes $\mathbf{b}_{i} \in B = \{\mathbf{b}_{i}\}_{i=1}^{N}$ for each of the $N$ referenced objects. Each 3D bounding box $\mathbf{b}_{i} = (c_{i}, \mathbf{t}_{i}, \mathbf{s}_{i}, {o}_{i})$ consists of the following components: category name $c_{i}$, box center $\mathbf{t}_{i} = (x_{i}, y_{i}, z_{i}) \in \mathbb{R}^{3}$, box dimensions $\mathbf{s}_{i} = (w_{i}, h_{i}, d_{i}) \in \mathbb{R}^{3}$ and orientation ${o}_{i} \in \mathbb{R}$.

The textual description $\mathcal{Y}$ is comprised of multiple sentences, each specifying the position and orientation of a single object within a canonical coordinate system. The canonical system is chosen to represent the scene in a 2D perspective, to ensure that the descriptions closely resemble those produced by a human observer. $\mathcal{Y}$ is initially generated using a rule-based approach formulated in three steps as follows: 1) The canonical coordinate system is partitioned into a 3 $\times$ 3 grid, creating rectangular patches. 2) Each of the predicted objects is assigned to a patch based on the center position $\mathbf{t}_{i}$ and orientation ${o}_{i}$ of its bounding box within these patches. 3) Per-object sentences are then formulated according to which patch the object is assigned to and object's orientation, \eg for the object assigned to the top-left patch, with category name $c_{i}$ and 90 degrees orientation, we generate: \textit{``A $c_{i}$ is placed at the top-left corner of the room, with a perpendicular orientation.''} Given the relatively limited linguistic variation inherent in rule-based descriptions, initial textual templates are provided to an LLM to serve as a basis for paraphrasing.

LayoutGPT \cite{layoutgpt} focused on improving an LLM's interpretation of the spatial knowledge by using a structured format featuring attribute-value pairs. We follow a similar strategy and adopt the standard CSS (Cascading Style Sheets) format to describe $\mathbf{b}_{i} $. Bounding box values are mapped to standard CSS format attributes and category name is employed as the selector.

\subsubsection{Prompt Construction.}
\label{subsubsec:promotconstruction} 

Recent progress in the field has showcased the impressive in-context learning capabilities of LLMs, which entail conditioning these models with both natural language instructions and a select set of task demonstrations. Following this conditioning, the LLM is then tasked with completing further instances based on the provided demonstrations. In line with this, our prompts to LLM include three main parts: task specifications, in-context exemplars and the query condition:
\begin{itemize}
\item \textbf{Task specifications:} Following prior work \cite{finetunedlanguagemodels, multitaskpromptedtraining} on enhancing LLMs instruction following capabilities, a task description is incorporated at the beginning of each prompt, which explains the goal of the task, establishes a standard for the 3D layout format in CSS style and provides unit information for attributes. Additionally, constraints are integrated into the prompts to guide the LLM and minimize errors (e.g., predicting overlapping boxes or placing objects out of bounds), during task completion.

\item \textbf{In-context learning:} Supporting exemplars for the in-context learning are selected by adopting the retrieval-based approach used in \cite{empiricalstudy, flamingo, layoutgpt}. When provided with a set of supporting exemplars $S = \{(\mathcal{C}_{m}^{s}, \mathbf{b}_{m}^{s})\}_{m=1}^{M}$
and the queried condition $\mathcal{C}_{q}$, the function $ f(\mathcal{C}_{k}^{s},\mathcal{C}_{q}) = \left\| rl_{k} - rl_{q}\right\|^{2} + \left\| rw_{k} - rw_{q}\right\|^{2}$ is computed between each element of the set and $\mathcal{C}_{q}$ following \cite{layoutgpt}, where $rl$ and $rw$ are the length and width of the scenes. Top-$k$ supporting exemplars with the shortest distance to $\mathcal{C}_{q}$ are selected for in-context learning, provided to LLM with the same format with $\mathcal{C}_{q}$.
\item \textbf{Query condition:} The inference condition $\mathcal{C}_{q}$, for which we want to predict the layout. 
\end{itemize}

\subsection{3D Scene Assembling from Layouts}
\label{subsec:3dsceneassemblingfromlayouts}

To assemble a scene based on the generated layouts, we rely on object retrieval. More specifically, for each of the bounding boxes, the nearest 3D model of the corresponding predicted category is retrieved from the furniture dataset based on the Euclidean distance of the bounding box dimensions as follows:

\begin{equation}
    d(\mathbf{s}_{i}, \mathbf{s}_{cand}) = \sqrt{(w_{i} - w_{cand}))^{2} + (h_{i} - h_{cand}))^{2} + (d_{i} - d_{cand}))^{2}},
\end{equation}
\noindent where $\mathbf{s}_{cand}$ denotes the bounding box dimensions for the candidate object from the dataset. The retrieved 3D model is placed within the scene according to the predicted box center $\mathbf{t}_{i}$ and orientation angle ${o}_{i}$.

\subsection{3D Scene Stylization}
\label{subsec:3dscenestylization}

Given a generated 3D scene, the users should be able to modify either the overall appearance of the scene or specific objects within it. Our approach facilitates scene editing by training a 3D Gaussian splatting model representing the generated scene and refining this representation according to the provided edit instruction, offering a flexible framework for 3D scene generation. 

\subsubsection{3D Gaussian Splatting Scene Representation.}
\label{sec:3dgaussianscenerepresentation}

3D Gaussian splatting (3DGS) \cite{gaussiansplatting} is a recent pioneering work for novel view generation, which represents the underlying scene as a collection of anisotropic 3D Gaussians defined by their center positions ${\mu} \in \mathbb{R}^3$ and 3D covariance matrices $\mathbf{\Sigma}$ parameterized as:

\begin{equation}
       \mathbf{{\Sigma}} = \mathbf{RSS^{\text{T}}R^{\text{T}}}.
\end{equation}

\noindent $\mathbf{R}$ denotes the rotation matrix and $\mathbf{S}$ is the scale matrix. Each 3D Gaussian is assigned a color $c$ represented with spherical harmonics (SH) coefficients, to capture the view-dependent appearance. To allow $\alpha$-blending of splats, Gaussians are associated with an opacity value $\alpha \in \mathbb{R}$. 

Unlike NeRF \cite{mildenhall2020nerf} which relies on volume rendering, 3DGS enables faster training and rendering through differentiable rasterization. A set of 3D Gaussians is rendered by projecting into the camera's image plane as 2D Gaussians, and assigned to individual image tiles. The color of each pixel $\mathbf{p}$ on the image plane is then determined as follows:

\begin{equation}
C(\mathbf{p}) = \sum_{i \in \mathcal{N}}^{} c_{i} \sigma_{i} \prod_{j=1}^{i-1} (1-\sigma_{j}), \: \: \: \:  \: \: \sigma_{i}= \alpha_{i} e^{-\frac{1}{2}(\mathbf{p}-\mu_{i})^{T} \mathrm{\Sigma}_{i}^{-1} (\mathbf{p}-\mu_{i})}
\end{equation}
where $\mathcal{N}$ denotes the Gaussians in this tile, $\sigma_{i}$ represents the influence of the Gaussian on the image pixel and $\mu_{i}$, $\mathbf{\Sigma}_{i}$, $c_{i}$, $\alpha_{i}$ are the position, covariance, color and opacity of the \textit{i}-th Gaussian respectively. For optimization, a weighted combination of $\mathcal{L}_{1}$ loss and SSIM loss is employed.

To initialize the centers of the 3D Gaussians, we utilize the mesh vertices of the retrieved objects and create a scene point cloud. This generated scene point cloud replaces the Structure-from-Motion (SfM) points used in the original work. To obtain the required training images for 3DGS optimization, we render the generated scene into RGB images $I$. We additionally render 2D  segmentation masks $\mathcal{M}$ to be used for 3D Gaussian scene editing.

\subsubsection{Scene Editing.}
\label{subsubsec:3dgaussiansceneediting}

For editing our 3DGS scenes based on the provided edit instruction, we build on top of Instruct-GS2GS \cite{igs2gs} which in turn extends Instruct-NeRF2NeRF \cite{inerf2nerf} for 3DGS scenes. Given the images used for training the 3DGS scene, Instruct-GS2GS updates the images individually using a diffusion model, namely Instruct-Pix2Pix \cite{instructpix2pix}. The diffusion model takes three inputs: 1) an unedited conditioning image $I_{0}^{v}$, a text instruction $c_{T}$ and the noisy version of the current render $I_{i}^{v}$ at optimization step $i$, where $v$ denotes the viewpoint from which the images are captured. Formally, the process of updating a single image is defined as:

\begin{equation}
I_{i+1}^{v} \leftarrow U_{\theta}(I_{i}^{v},t;I_{0}^{v},c_{T}),
\end{equation}

\noindent where $t$ is the noise level within the constant range $[t_{\text{min}},t_{\text{max}}]$, $U_{\theta}$ is the sampling process of DDIM \cite{stablediffusion} and $I_{i+1}^{v}$ is the edited image respecting the text instruction $c_{T}$ and the unedited conditioning image $I_{0}^{v}$. The training is performed by editing all the training images during a dataset update at every 2.5k training iterations.

Despite achieving general scene-level stylization, \cite{igs2gs} lacks of support for more fine-grained object-level editing. To enable object-level edits, we use the binary masks $m_{\{o\}_{k=1}^{K}}$. Each $m_{\{o\}_{k=1}^{K}}$ is obtained by binarizing 2D segmentation masks $\mathcal{M}$ for the set of objects to edit ${\{o\}_{k=1}^{K}}$. The set ${\{o\}_{k=1}^{K}}$ is defined by extracting the referenced category names from the text instruction $c_{T}$. Having the unedited conditioning image $I_{0}^{v}$, the edited image $I_{i+1}^{v}$, and the binary mask $m_{\{o\}_{k=1}^{K}}$, we keep only the edits at the pixels of the target object set:

\begin{equation}
I_{i+1}^{v} = m_{\{o\}_{k=1}^{K}} \odot I_{i+1}^{v} + (1 - m_{\{o\}_{k=1}^{K}}) \odot I_{0}^{v},
\end{equation}

\noindent where $\odot$ denotes element-wise multiplication of the image pixels. This way, the other pixels within $I_{i+1}^{v}$ are set to their unedited versions, enabling an object-level editing of training images. While training the 3DGS scene with the edited images, the mask $m_{\{o\}_{k=1}^{K}}$ is also applied for $\mathcal{L}_{1}$ and $\text{SSIM}$ losses, ensuring gradient propagation only to the target objects.

\section{Experiments}
\label{sec:experiments}

\subsubsection{Datasets.}
\label{subsubsec:implementationdetails}

To create a set of supporting exemplars for the layout generation, we employ the 3D-FRONT dataset \cite{3dfront1}, a large-scale indoor scene dataset with 18,797 rooms furnished with 3D textured objects from 3D-FUTURE \cite{3dfront2}. Following the same pre-processing steps as \cite{atiss} and excluding non-rectangular floor plans from the test set similar to \cite{layoutgpt}, we obtained 3397 train, 453 validation, 423 test bedroom and 690 train, 98 validation, 53 test living room scenes. During 3D scene assembling, 3D models are retrieved from the 3D-FUTURE dataset.

\subsubsection{Implementation Details.}
\label{subsubsec:implementationdetails}

We use ChatCompletions API \footnote{See \href{https://platform.openai.com/docs/api-reference/chat}{https://platform.openai.com/docs/api-reference/chat}}, which enables a series of dialogue exchanges between the user and the LLM, to query GPT-4 \cite{gpt4}. BlenderProc \footnote{See \href{https://github.com/DLR-RM/BlenderProc}{https://github.com/DLR-RM/BlenderProc}} is employed to render 250 images per assembled scene with 512 $\times$ 512 resolution. After relocating the center of each scene to the origin of the coordinate system, we sample camera positions on the upper hemisphere at a distance of 1.5$r$ from the center where $r$ is the diagonal length of the room, with an elevation angle of 35 degrees. We use the Splatfacto model from NeRFstudio \cite{nerfstudio} for 3DGS training, by optimizing each scene for a maximum of 20k iterations. For editing the 3DGS training images, we use InstructPix2Pix \cite{instructpix2pix} as the diffusion model, with classifer guidance scales $s_{T} = [3.5, 12.5]$ for the text and $s_{I} = 1.5$ for the image. The dataset updates are performed every 2.5k iterations following Instruct-GS2GS \cite{igs2gs}. The InstructGS2GS training takes approximately 15-20 minutes per scene on a single A6000, most of it spent on diffusion model sampling. The rendering speed is 101.21 fps.

\subsection{3D Layout Generation}
\label{subsec:3Dlayoutgenerationeval}

\subsubsection{Evaluation Setup and Baselines.} In this section we evaluate the first stage of our approach, therefore we only focus on assessing the generated layouts. To our knowledge, \textit{SceneTeller} is the first approach that conducts conditional layout generation guided by text prompts to control the position and orientation of every object, targeting a user-friendly framework for novices. Hence, replicating an interaction mode that is natural for humans is pivotal for our framework. Based on this, our layout generation performance is compared with the performance of \textit{humans} solving the same task. In particular, each participant of our user study is presented with a series of textual descriptions, each accompanied by an empty layout according to the room dimensions. For each text prompt, participants are requested to draw bounding boxes and label them with the object names. Each text prompt is given to at least 2, at most 3 participants. This study is conducted with 12 participants and 40 text prompts. To generate text prompts, 40 layouts are selected from the test set and their text prompts are generated by providing the initially rule-based formed sentences to GPT for paraphrasing, as described in \cref{subsec:languagedriven3dlayoutgeneration}.

\subsubsection{Metrics.} Based on the predicted number of bounding boxes in each category, we report recall, precision and accuracy \cite{layoutgpt}. An accuracy score of 100 indicates that the number of bounding boxes and their corresponding categories precisely match the ground-truth annotations. Otherwise, a score of 0 is assigned to the scene. All values are in percentage (\%). We additionally report mean Intersection over Union (mIoU) between the predicted and ground-truth bounding boxes and out-of-bound rates, indicating the percentage of scenes where furniture extends beyond the floor plan boundary. We refer to the supplementary material for detailed metric descriptions.

\subsubsection{Results.}

\cref{tab:layoutgenerationeval} presents the quantitative evaluation results of 3D layout generation. Our in-context-learning-based 3D layout generation module performs better than human participants in terms of mIoU, showcasing the effectiveness of LLMs' visual reasoning capabilities. In-context exemplars further enhance this performance by aiding in reasoning about the current data distribution. However, human participants possess their own understanding of the physical world, leading to variations in their notions of object sizes and prompt reasoning, resulting in slightly poorer performance in this assessment. Both our layout generation module and human participants performed excellently in terms of predicting the number of bounding boxes per category. Humans sometimes forget or miss objects mentioned in the prompt, which explains their slightly imperfect recall score. The ground-truth layouts from the dataset involve objects slightly exceeding the floor plan boundary, leading to 15.15\% OOB rate. Human participants are able to place the objects within the layout more accurately.

\begin{table}[h]
  \caption{\textbf{Quantitative evaluation of 3D layout generation.} The proposed in-context learning based 3D layout generation module performs better than human participants in terms of predicting object sizes, enhanced by in-context exemplars aiding in reasoning about the current data distribution. All values are in percentage (\%).} 
  \label{tab:layoutgenerationeval}
  \centering
  \begin{tabular}{c|c|ccc|c}
    \toprule
     \textbf{Method} & OOB $\downarrow$& rec.$\uparrow$ & prec.$\uparrow$ & acc.$\uparrow$ & mIoU $\uparrow$\\
    \midrule
    Ours &  56.78 & 100.00 & 100.00 & 100.00 & 44.60  \\
    Human  & 12.09 & 98.79 & 100.00 & 95.00 &  28.70  \\
    GT & 15.15 & 100.00 & 100.00 & 100.00 & 100.00   \\
  \bottomrule
  \end{tabular}
\end{table}

\subsection{3D Scene Generation}
\label{subsec:3Dlayoutgenerationeval}

\subsubsection{Baselines.} Our approach is evaluated and compared with three state-of-the-art 3D scene generation methods. Set-the-scene \cite{setthescene} employs compositional NeRFs that are initialized based on user-defined object proxies to have more control over individual objects. To compare with \cite{setthescene}, layouts generated by our method are provided as input to their pipeline. GSGEN \cite{gsgen} is a 3DGS-based approach, following a two-stage pipeline for geometry optimization and appearance refinement. LucidDreamer \cite{luciddreamer} employs render-refine-repeat paradigm by iteratively warping and inpainting the previously generated image to create an initial point cloud, which is then used for 3DGS scene training. 

\begin{figure}[!htbp]
  \centering
   \includegraphics[height=14.9cm]{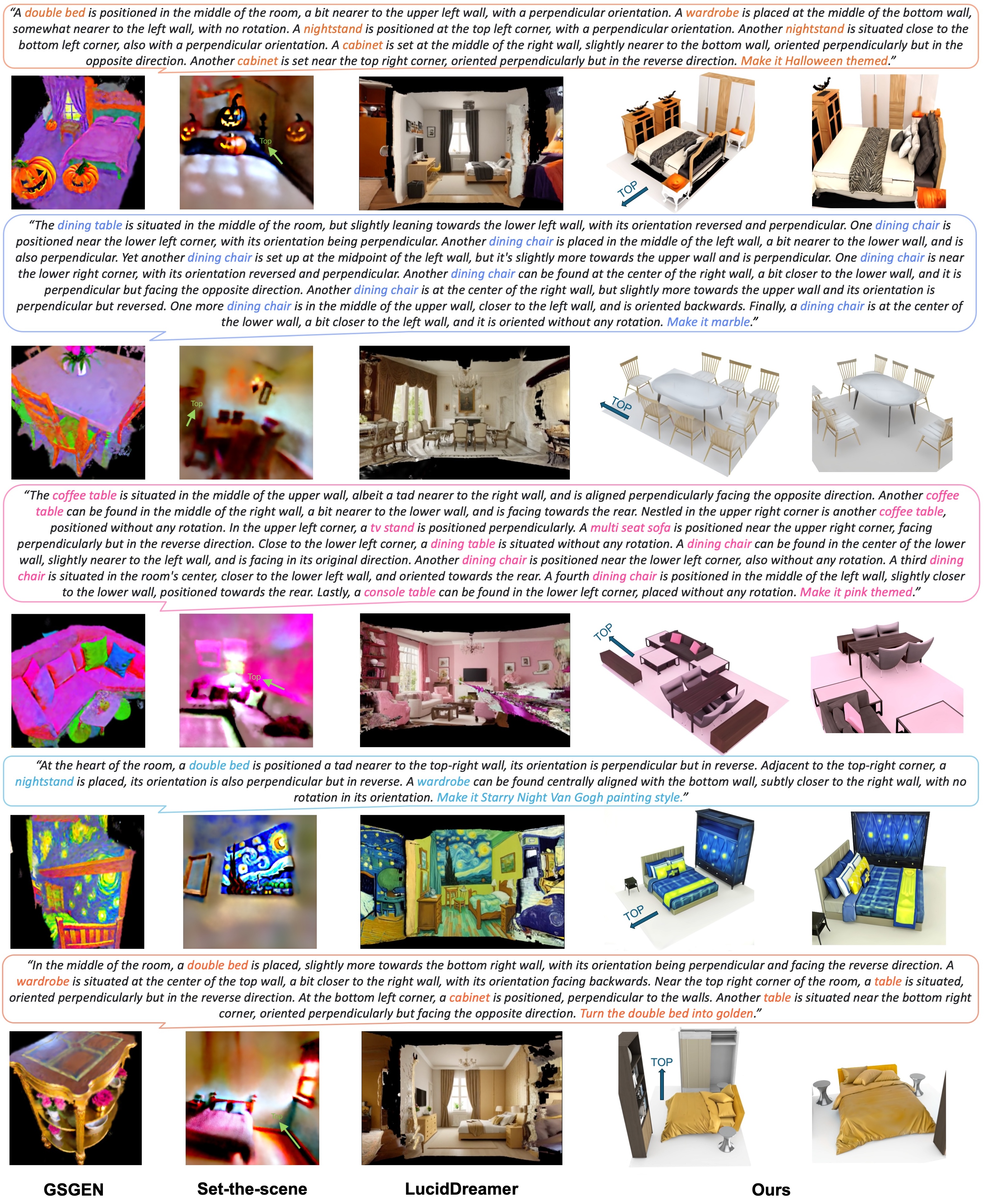}
   \caption{\textbf{Qualitative comparison with state-of-the-art text-to-3D scene generation methods}. The top/upper walls are marked with arrows for your reference, if this information is available within the generations. \textit{SceneTeller} is able to generate high-quality scenes, with superior geometric fidelity and 3D consistency.}
   \label{fig:qualeval}
\end{figure}

\subsubsection{User Study.}

We conduct a user study with 30 participants to assess the performance of our method in generating high-quality text-consistent realistic 3D scenes. Human evaluators are presented with 10 scenes, each accompanied by a textual prompt describing its layout and style. For each scene, they are requested to rate it on a scale from 1 to 5 (5 being the best) based on four criteria: (a) Realism, (b) Text Alignment, (c) Geometric Fidelity and (d) Compositional Plausibility.

We present the results of the study in \cref{tab:userstudy}. Our method consistently outperforms the existing state-of-the-art 3D scene generation methods across all the four criteria by a significant margin, showcasing the merits of \textit{SceneTeller} in generating realistic high-quality 3D scenes solely from text prompts. For additional details of the study and extensive analysis of the results, we refer to the supplementary material.

\begin{table}[h]
  \centering
  \caption{\textbf{User study results}. Study conducted with 30 participants to assess the performance of state-of-the-art methods in generating 3D scenes based on four criteria with a scale from 1 to 5 (5 being the best and vice versa). Our method consistently outperforms the existing state-of-the-art 3D scene generation methods across all the four criteria by a significant margin, showcasing the merits of \textit{SceneTeller} in generating realistic high-quality 3D scenes solely from text prompts. }
  \label{tab:userstudy}
  \makebox[\textwidth][c]
    {
    \begin{tabular}{lccccc}
    \toprule
    \multicolumn{1}{l}{\textbf{Method}} & \multicolumn{1}{c}{\textbf{Input Type}}  & \multicolumn{4}{c}{\textbf{User Study}}  \\

    \cmidrule{3-6}
    & & Realism & \makecell{Text \\ Alignment} & \makecell{Geometric \\ Fidelity} & \makecell{Compositional \\ Plausibility}  \\
    
    \hline \hline
    GSGEN \cite{gsgen} & T & 1.313 & 1.375 & 1.439 & 1.392 \\
    LucidDreamer \cite{luciddreamer} & T & 3.543 & 2.751 & 3.050 & 3.116 \\
    Set-the-scene \cite{setthescene} & L + T & 1.963 & 1.900 & 2.112 & 2.181 \\
    \hline
    Ours & T & \textbf{4.154} & \textbf{4.189} & \textbf{4.479} & \textbf{4.323} \\
    \bottomrule
    \end{tabular}
    }
\end{table}

\subsubsection{Quantitative Evaluation of Stylization.} To evaluate the consistency of the generated scene style with textual descriptions, we utilize the CLIP Score \cite{clipscore} as the metric. Given that the CLIP score cannot reason about the parts of the prompts describing the layout, we only provide style information to CLIP in the format: \textit{'A Barbie style bedroom'} or \textit{'A bedroom with Barbie style nightstands'} for all methods. As shown in \cref{tab:styleeval}, our method consistently generates scenes adhering to the given style description. 

\begin{table}[h]
  \centering
  \caption{\textbf{Quantitative comparison of stylization.} We report the CLIP Score to evaluate the consistency of the generated scene style with textual descriptions. Our method consistently generates scenes adhering to the given style description.}
  \label{tab:styleeval}
    \begin{tabular}{l|ccc|c}
    \toprule
    \multicolumn{1}{c|}{}  &  \multicolumn{1}{c}{GSGEN \cite{gsgen}} & \multicolumn{1}{c}{LucidDreamer \cite{luciddreamer}} & \multicolumn{1}{c}{Set-the-scene \cite{setthescene}}  & \multicolumn{1}{|c}{Ours}   \\  
    \hline\hline
    CLIP Score $\uparrow$ & 26.478 & 30.715 & 27.188 & \textbf{30.923} \\
    \bottomrule
    \end{tabular}
\end{table}

\subsubsection{Qualitative Evaluation.}
Visual comparison of \textit{SceneTeller} with state-of-the-art text-to-3D methods are provided in \cref{fig:qualeval}. \textit{SceneTeller} yields high-quality 3D scene generations, surpassing the baselines in terms of text-alignment and 3D consistency by predicting an initial 3D layout conditioned on the textual descriptions. Unlike some of the state-of-the-art baselines producing unrealistic appearance (\eg comic or cartoon-like characteristics),  \textit{SceneTeller} is able to produce more realistic object shapes and appearances.

\subsubsection{Realism of the Generated Scenes.} We evaluate all methods using FID score with a real indoor dataset ScanNet \cite{scannet} on 10 generated scenes by rendering 50 images from each of them. As shown in \cref{tab:FIDresults}, our method consistently generates more realistic scenes.

\begin{table}[h]
\caption{FID score with the ScanNet dataset \cite{scannet}.}
\centering
\begin{tabular}{l|ccc|c}
\toprule
\multicolumn{1}{c|}{}  &  \multicolumn{1}{c}{GSGEN \cite{gsgen}} & \multicolumn{1}{c}{LucidDreamer \cite{luciddreamer}} & \multicolumn{1}{c}{Set-the-scene \cite{setthescene}}  & \multicolumn{1}{|c}{Ours}   \\
\hline \hline   
FID $\downarrow$ & 297.10  & 238.75 & 297.57 & \textbf{167.27}    \\
\bottomrule
\end{tabular}
\label{tab:FIDresults}
\end{table}

\subsubsection{Applicability to Real-world.} 
While current design tools enable professionals to create floor plans, visualize 3D models, and render photo-realistic images, they are not suitable for novice users unfamiliar with such software tools. By enabling users to interact solely through plain natural language prompts, our approach also empowers novice users to flexibly design their own rooms.

\subsection{Ablation Study}

\subsubsection{Effect of design choices for 3D layout generation.} To assess the effectiveness of individual components in our design, we conduct an ablation study. Specifically, we examine two aspects: 1) the prompt source for in-context exemplars, and 2) the technique for selecting in-context exemplars. Regarding the \textbf{prompt source}, we present exemplars where the textual description is generated either solely by the rule-based formulation (R) or by a combination of the rule-based formulation and GPT paraphrasing (R+GPT) to introduce linguistic variation as described in \cref{subsec:languagedriven3dlayoutgeneration}. For the \textbf{exemplar selection} technique, we experiment with a \textit{random} and a \textit{retrieval}-based exemplar selection following \cite{layoutgpt}, $k=8$ exemplars. Additionally, we experiment with a third set of exemplars, denoted as \textit{pos+neg}, where 4 exemplars are selected by retrieval, and the other 4 are chosen from negative exemplars involving out-of-bound (OOB) or overlapping objects. The negative ones are from the filtered samples during the pre-processing of the dataset, and the LLM is specifically instructed to avoid them. \cref{tab:ablationlayoutdesignchoices} verifies that providing in-context exemplars with linguistic variation (R+GPT) selected with a \textit{retrieval}-based approach performs the best, especially in terms of OOB rate by constraining the in-context learning of LLM to similar layouts. Using the \textit{pos+neg} set also considerably improves the performance in terms of mIoU.

\begin{table}[h]
  \centering
  \caption{\textbf{Effect of design choices for 3D layout generation.} We report the 3D layout generation performance w.r.t different design choices such as the prompt source and the exemplar selection. Providing in-context exemplars to LLM with linguistic variation (R+GPT) selected with a \textit{retrieval}-based approach, performs the best.}
  \label{tab:ablationlayoutdesignchoices}
  \makebox[\textwidth][c]
    {
    \begin{tabular}{lccccccccc}
    \toprule
    \multicolumn{1}{l}{\textbf{}}  & \multicolumn{1}{p{2cm}}{\centering \textbf{Prompt} \\ \textbf{source}}  & \multicolumn{3}{p{3cm}}{\centering \textbf{Exemplar} \\ \textbf{Selection}} & \multicolumn{1}{l}{\textbf{}} & \multicolumn{3}{p{3.0cm}}{\centering \textbf{Numerical} \\ \textbf{Reasoning}} & \multicolumn{1}{l}{\textbf{Bbox}}  \\

    \cmidrule{3-5}
    \cmidrule{7-9} 
    &  & random & retrieval & pos+neg & OOB(\%) $\downarrow$ & rec.$\uparrow$ & prec.$\uparrow$ & acc.$\uparrow$ & mIoU $\uparrow$\\
    
    \hline \hline
    \fontfamily{pcr}\selectfont 1 & R & \checkmark & & & 74.12 & 100.00 & 100.00 & 100.00 & 31.10 \\
    \fontfamily{pcr}\selectfont 2  & R &  & \checkmark & & 64.87 & 100.00 & 100.00 & 100.00 & \textbf{45.50} \\
    \fontfamily{pcr}\selectfont 3  & R &  &  & \checkmark  & 74.10 & 100.00 & 100.00 & 100.00 & 41.90 \\
    \fontfamily{pcr}\selectfont 4  & R + GPT & \checkmark & & & 76.11 & 99.10 & 99.10 & 99.10 & 29.90  \\
    \fontfamily{pcr}\selectfont 5  & R + GPT &  & \checkmark & & \textbf{54.65} & 100.00 & 100.00 & 100.00 & 44.20 \\
    \fontfamily{pcr}\selectfont 6  & R + GPT &  &  & \checkmark &  66.45 & 100.00 & 100.00 & 100.00 & 43.90 \\
    \bottomrule
    \end{tabular}
    }
\end{table}

\section{Conclusion}
\label{sec:conclusion}

In this paper, we presented a novel pipeline for 3D room generation from a textual description. Our approach supports style editing either on the individual object or on the entire scene level. Rooms generated by our method were preferred by human evaluators over the baseline results almost in all cases, confirming its efficacy on this challenging task. Importantly, our approach can be used by novices, thus making a step towards democratizing 3D scene design.


\section*{Acknowledgements}
We thank members of the Bosch-UvA Delta Lab and anonymous reviewers for helpful discussions and feedback. This project was generously supported by the Bosch Center for Artificial Intelligence.

%
%
\bibliographystyle{splncs04}
\bibliography{main}
\end{document}